\pdfoutput=1
\documentclass[11pt]{article}
\usepackage{booktabs}
\usepackage{graphicx} 
\usepackage{algpseudocode}
\usepackage{hyperref}
\usepackage[]{acl}
\usepackage{multirow}
\usepackage[inline]{enumitem}
\usepackage{enumerate}  
\usepackage{amsfonts}
\usepackage{latexsym}
\usepackage{xspace}
\usepackage{amsmath}
\usepackage[T1]{fontenc}
\usepackage{pifont}
\usepackage{amssymb}
\usepackage{newfloat}
\usepackage{listings}
\usepackage{multirow}
\usepackage{booktabs}
\usepackage{times}
\usepackage[inline]{enumitem}
\usepackage{latexsym}
\usepackage[T1]{fontenc}    
\usepackage[utf8]{inputenc}
\usepackage[inline]{enumitem}
\usepackage[dvipsnames]{colortbl}
\usepackage[utf8]{inputenc}
\usepackage[T1]{fontenc}
\usepackage{babel} 
\usepackage[toc,page]{appendix}
\usepackage{makecell}
\usepackage{soul}
\usepackage{verbatim}
\usepackage{arydshln}
\usepackage{amssymb}
\usepackage{tcolorbox}
\usepackage{xcolor}
\usepackage{color}
\usepackage{colortbl}

\usepackage{graphicx} 
\usepackage[linesnumbered,ruled,vlined]{algorithm2e}
\usepackage{adjustbox}
\usepackage{footmisc}
\usepackage{microtype}
\usepackage{bbm}
\usepackage{wasysym}
\usepackage{tikz}
\usepackage{tcolorbox}
\usepackage{siunitx}
\usepackage{pifont}
\usepackage{longtable}
\usepackage{supertabular}
\usepackage[fixed]{fontawesome5}

\usepackage[utf8]{inputenc}

\usepackage{microtype}
\usepackage[nameinlink]{cleveref}
\usepackage{inconsolata}

\title{ Learning to Use Tools via
Cooperative and Interactive Agents \\with Large Language Models}
\definecolor{Gainsboro}{rgb}{0.86, 0.86, 0.86}

\definecolor{Gray}{gray}{0.95}
\definecolor{LightCyan}{rgb}{0.88,1,1}

\newcommand{\ie}{\emph{i.e.,}\xspace}
\newcommand{\eg}{\emph{e.g.,}\xspace}
\newcommand{\code}[1]{{\ttfamily#1}}

\newcommand{\ours}{ConAgents\xspace}
\newcommand{\first}{grounding agent\xspace}
\newcommand{\second}{execution agent\xspace}
\newcommand{\third}{review agent\xspace}

\newcommand{\distill}{SPAN\xspace}

\newtcbox{\hlprimarytab}{on line, rounded corners, box align=base, colback=c3!10,colframe=white,size=fbox,arc=3pt, before upper=\strut, top=-2pt, bottom=-4pt, left=-2pt, right=-2pt, boxrule=0pt}
\newtcbox{\hlsecondarytab}{on line, box align=base, colback=red!10,colframe=white,size=fbox,arc=3pt, before upper=\strut, top=-2pt, bottom=-4pt, left=-2pt, right=-2pt, boxrule=0pt}
\newcommand{\daugshifted}{\raisebox{0.5\depth}}

\newcommand{\daulg}[1]{{\hlsecondarytab{\daugshifted{#1}}}}

\newcommand{\ga}{$\mathcal{M}_G$\xspace}
\newcommand{\ea}{$\mathcal{M}_E$\xspace}
\newcommand{\oa}{$\mathcal{M}_R$\xspace}

\author{
Zhengliang Shi$^{1}$\space\space\space
Shen Gao$^{2}$\space\space\space
Xiuyi Chen$^3$\space\space\space
Yue Feng$^4$\space\space\space
Lingyong Yan$^3$\space\space\space
\\
{\bf
    Haibo Shi$^3$\space\space\space
    Dawei Yin$^3$\space\space\space
    Pengjie Ren$^1$\space\space\space
    Suzan Verberne$^5$\space\space\space
    Zhaochun Ren$^5$
}\\  
$^1$Shandong University~~$^2$University of Electronic Science and Technology of China \\
$^3$Baidu Inc., Beijing, China~~$^4$University of Birmingham, Birmingham, UK\\
$^5$Leiden University, Leiden, The Netherlands \\
shizhl@mail.sdu.edu.cn~~z.ren@liacs.leidenuniv.nl
}
\begin{document}
\maketitle
\begin{abstract}

Tool learning empowers large language models (LLMs) as agents to use external tools and extend their utility.
Existing methods employ one single LLM-based agent to iteratively select and execute tools, thereafter incorporating execution results into the next action prediction.
Despite their progress, these methods suffer from performance degradation when addressing practical tasks due to: (1) the pre-defined pipeline with restricted flexibility to calibrate incorrect actions, and (2) the struggle to adapt a general LLM-based agent to perform a variety of specialized actions.
To mitigate these problems, we propose \ours, a \textbf{Co}operative and i\textbf{n}teractive \textbf{Agents} framework, which coordinates three specialized agents for tool selection, tool execution, and action calibration separately.
\ours introduces two communication protocols to enable the flexible cooperation of agents.
To effectively generalize the \ours into open-source models, we also propose specialized action distillation, enhancing their ability to perform specialized actions in our framework.
Our extensive experiments on three datasets show that the LLMs, when equipped with the \ours, outperform baselines with substantial improvement (\ie up to 14\% higher success rate).

\end{abstract}

\section{Introduction}\label{sec:intro}

\begin{figure}[t]
        \centering
	\includegraphics[width=1\linewidth]{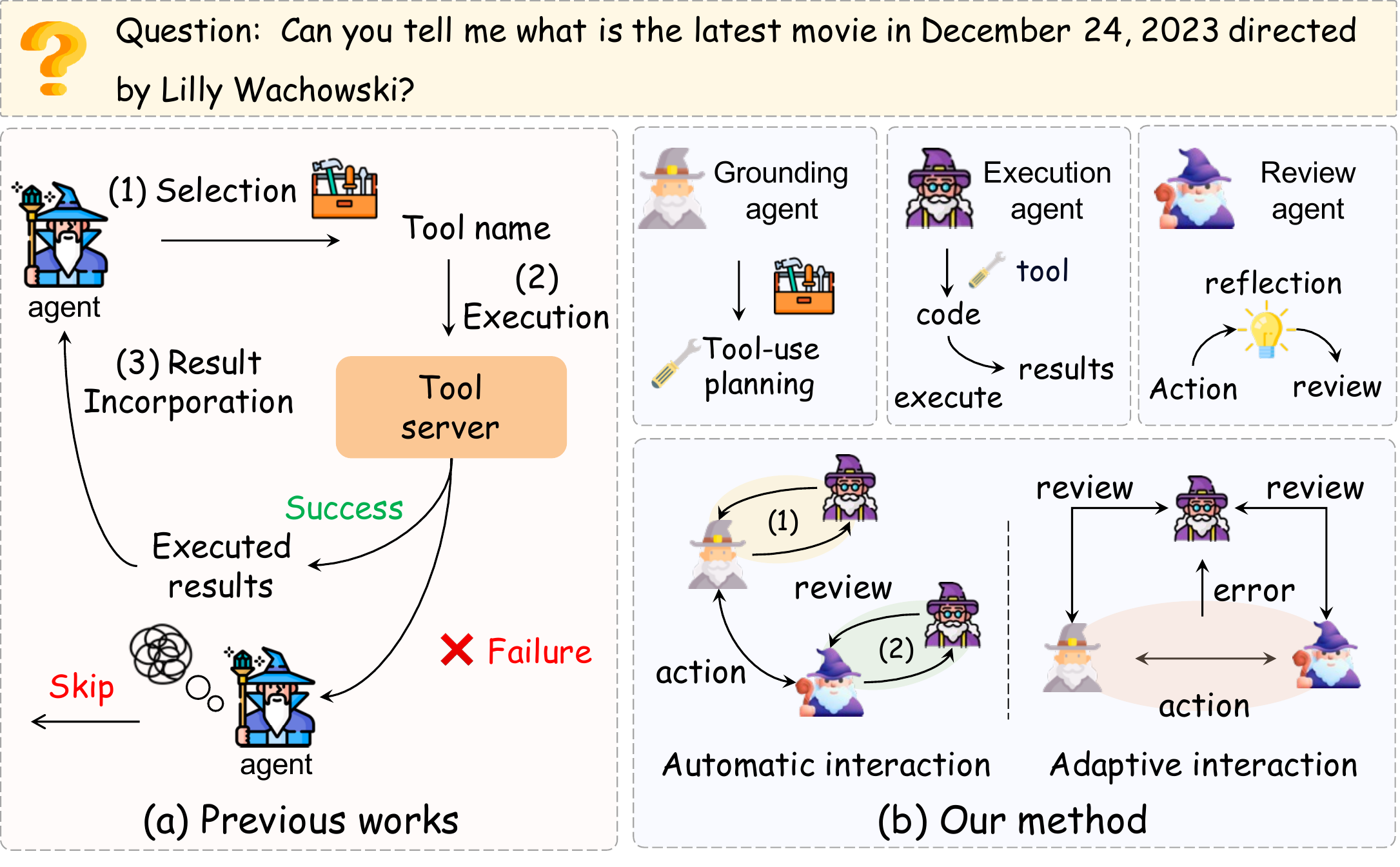}
        \caption{Comparison between (a) existing single-agent tool learning method and (b) our cooperative agent framework \ours.
        The \ours coordinates three agents through two proposed communication protocols, \eg automatic and adaptive interaction.}
 \label{fig:intro}
\end{figure}

Although large language models (LLMs) have achieved remarkable performance in a broad range of natural language processing tasks~\citep{self-instruction,chang2023survey}, they still encounter inherent limitations such as out-of-date information~\citep{toolw,mallen2023not}.
\textbf{Tool learning} is proposed to equip LLMs with various auxiliary resources, \eg a search engine~\citep{webcpm,webgpt} or a calculator~\citep{toolformer,pal}, which 
empower them as tool-use agents and improve their proficiency in tackling concrete complex tasks.
As shown in Figure~\ref{fig:intro}(a), most previous studies allow the LLM-based agent to interleave multiple actions in a pre-defined order to interact with tools~\cite{react, yang2023mm,zhuang2023toolqa}.
The agent first breaks down the task and \textbf{\texttt{plans}} a series of tools in a step-by-step manner.
For each step, the agent \textbf{\texttt{executes}} the tools by passing arguments and continuously \textbf{\texttt{incorporates}} useful intermediates into the next action prediction.

Despite the advancement of existing methods, they face two challenges in practice.
\textbf{First},  most of them alternate the planning and execution with a pre-defined pipeline~\cite{yang2023mm,restgpt},  which \textit{inevitably constrains their flexibility in handling exceptional errors} that frequently occur during a tool-use workflow~\citep{shi2024chain,wang2023mint,prasad2023adapt}.
When failing to invoke tools, it is crucial to enable agents to revise their incorrect actions instead of directly shifting to the next step with the error response of previous steps.
\textbf{Second}, \textit{it is struggling to adapt a single LLM-based agent to learn a variety of specialized actions in solving a task}~\citep{dziri2023faith,yin2023lumos}.
Solving a practical task involves varied actions with substantial differences, \eg planning, execution, and reflection, drawing upon different facets of the LLMs~\cite{shen2024small,qiao2024autoact}.
Therefore, developing effective agent flow and adapting tool-use models to solve practical tasks remains a challenging research topic.

In this work, we propose \ours, a \textbf{Co}operative and i\textbf{n}terative \textbf{Agents} framework for tool learning tasks.
As illustrated in Figure~\ref{fig:intro}, \ours decomposes the overall tool-use workflow using three specialized agents: \textit{Grounding}, \textit{Execution}, and \textit{Review} agents.
The \first reasons the task description and grounds it into planning by specifying which tool to use.
The \second follows the planning to execute the selected tool by generating executable code.
The \third reviews the incorrectness in planning or execution, providing feedback for revision.
To enable the dynamic cooperation of these specialized agents, we propose two communication protocols, including automatic and adaptive interaction.
In the process of \textit{automatic interaction}, the \third provides real-time reviews to calibrate incorrect actions. 
Thus, the agent flow alternates between the planning-review and execution-review phases as shown in Figure~\ref{fig:intro}.
In the process of \textit{adaptive interaction}, the \third only provides feedback when exceptional errors are captured while executing the tools.

For a comprehensive evaluation, we conduct experiments on two benchmarks, \ie ToolBench and RestBench, using various LLMs as backbones.
We find that \ours outperforms the state-of-the-art baseline with both communication protocols (6\% improvement in Success Rate on average).

Despite closed-source LLMs performing well with our framework, we find the open-source models may struggle with the modulized agent flow. 
Thus, we propose an approach called \textbf{sp}ecialized \textbf{a}ction distillatio\textbf{n} (\distill), enhancing the performance of open-source models in \ours.
We heuristically sample 2,919 high-quality tasks from the ToolBench~\cite{toolllm} training set, and cluster them based on their similarity, retaining only one task in each cluster to avoid  duplication.
For each task, we guide the GPT-4 to generate solutions using \ours, and reorganize them into actions tailored to specialized agent functionalities in \ours.
These actions are separately distilled into different student models through instruction tuning.
We employ parameter-efficient tuning techniques, \ie LoRA~\cite{hu2021lora}, further extending our distillation method into low-resource scenarios.
Experiment results show that our distillation method empowers open-source models with strong performance with only 500 training examples.

Our contributions are summarized as follows:
(1) We propose \ours, a cooperative and interactive agents framework, for tool learning tasks. 
\ours coordinates three specialized agents with two communication protocols to solve a complex task.
(2) We propose specialized action distillation (\distill), which more effectively enables open-source models to work with the \ours;
(3) Both automatic and human evaluation conducted on two benchmarks validate the superiority of \ours.

\section{Related Work}\label{sec:related-work}

\paragraph{LLMs for tool learning.}
Enhancing LLMs with external tools has been proven a promising method for solving practical tasks~\cite{chemcrow,qu2024tool,wang2024tools}. 
Previous works empower a tool-learning agent typically by supervised fine-tuning~\citep{patil2023gorilla,gpt4tools,gao2023confucius} or prompt learning~\citep{chameleon,huggingfacegpt}.
Specifically, the former trains LLMs on tool-use dataset~\cite{self-instruction}, teaching LLMs how to use tools from the data.
The latter directly demonstrates tool usages to LLM using in-context examples ~\citep{art,rci}.
However, solving complex tasks with tools involves various actions, \eg deciding which tools to use, what arguments to pass, and how to utilize the results~\citep{toolformer,qiao2024autoact}.
Compelling one single agent to learn all abilities places even greater pressure on it~\citep{yin2023lumos,prasad2023adapt}.
In addition, as the tasks become complex, LLMs-based agents struggle to incorporate lengthy task-solving contexts to predict the next actions correctly due to their limited working memory~\citep{pmlr-v202-shi23a}.
In contrast, our proposed \ours coordinates three specialized agents, generating a solution through agent cooperation.

\paragraph{Multi-agent cooperation.}
Synergizing multiple agents has demonstrated strong performance on a variety of tasks~\citep{liu2023dynamic, sun2023corex, zhang2023agentcf}, enhancing the capabilities of individual agents~\citep{talebirad2023multi,mohtashami2023social,qian2023communicative}.
Recent studies take multiple agents into a debate for a fixed number of rounds~\citep{wang2023mac,liang2023encouraging},  boosting their factuality~\citep{cohen2023lm} and reasoning abilities~\citep{du2023improving,fu2023improving}.
In the tool learning tasks, recent work separately implements the task planning and execution with different agents, thereby reducing the workload of a single agent~\cite{shen2024small, restgpt, qiao2024autoact}.
Despite their progress, their agent flow is simplified into a pre-defined pipeline~\citep{prasad2023adapt}, struggling to handle exceptional errors that frequently occur during the tool-use workflows~\citep{zhuang2023toolqa,wang2023mint}.
In our work, we propose two communication protocols, which enable the action calibrations and dynamic cooperation of agents.

\section{Methodology}\label{sec:method}
\begin{figure*}[htbp]
        \centering
	\includegraphics[width=1
 \linewidth]{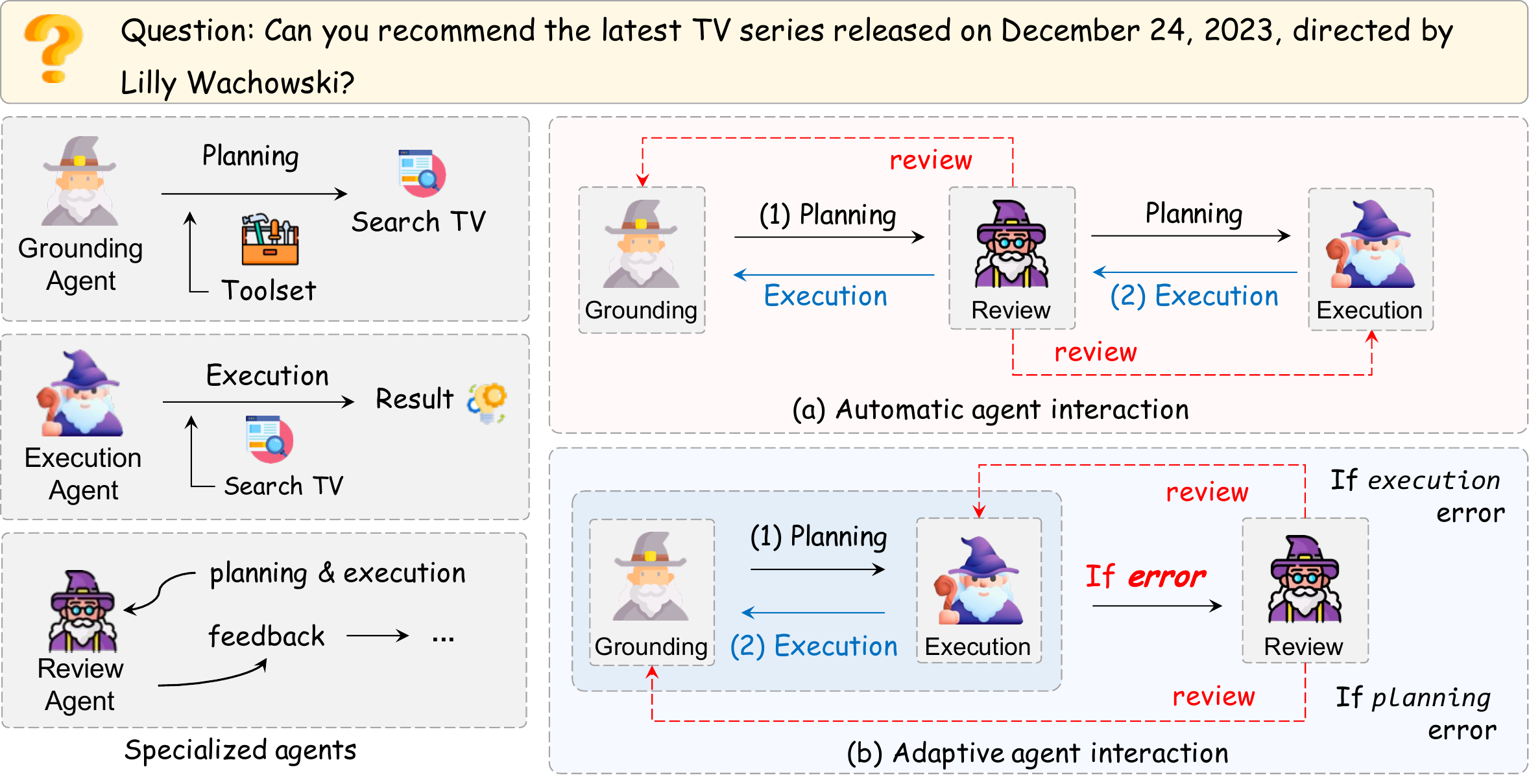}
        \caption{Our proposed cooperative and interactive agent framework. 
The \textbf{left} shows the three specialized agents in our framework (\S~\ref{sec:framework}).
The \textbf{right} illustrates  two proposed communication protocols to coordinate these specialized agents, including the \textit{automatic} and \textit{adaptive} communication (\S~\ref{sec:protocol}).
        }
 \label{fig:method}
\end{figure*}

\subsection{Overall Framework}\label{sec:framework}

Our cooperative framework, \ours, is proposed to enable the dynamic cooperation of agents to solve complex tasks.
As shown in Figure~\ref{fig:method}, \ours streamlines and modularizes the workflow of tool learning tasks into a \first \ga,  \second \ea, and \third \oa.
These agents are implemented with different system prompt or learnable parameters.
Given a complex task $x$, the \ga first decomposes $x$ into simpler sub-tasks and generates tool-use planning $t$ in a step-by-step manner.
For each step, the \ea executes the selected tool by writing executable code following the planning $t$.
The execution result $r$ is then incorporated into the context of the \first \ga to predict planning in the next iteration.
The \oa is employed to simulate an expert to provide feedback to agent \ga and \ea, guiding them to revise their incorrect planning or execution. 
To coordinate these three specialized agents, we explore and analyze two communication protocols, including the \textit{automatic} and \textit{adaptive} interactions.

\subsection{Specialized Agents}
\paragraph{Grounding Agent.}\label{sec:grounding}
The \first is designed to break down an input task and generate a series of tool-use planing.
At $i$th iteration, the \first generates planning $t_i$  on the condition of the task $x$ and current trajectory $\mathcal{H}_i =\{(t_j, r_j) | j < i\}$, consisting of the accumulation of previous planning $t_{<i}$ and results $r_{<i}$.
It can be formulated as:
\begin{equation}
    t_i=\mathcal{M}_G(x, \mathcal{S}, \mathcal{H}_i), 
\end{equation}
where $t_i$ contains a tool selected from the provided toolset $S$ and necessary arguments to invoke the tool, such as \texttt{``Use the Bing search to find a movie shown on Dec 24, 2023''}.

\paragraph{Execution Agent.}\label{sec:execution}
Following the generated planning $t_i$, the \second \ea executes the selected tool by generating executable code $c$ with the assistance of the tool documentation $d$.
This process can be formulated as:
\begin{equation*}
    \begin{aligned}
    c_i &= \mathcal{M}_E (d, t_i).
    \end{aligned}
    \label{eq:arguments}
\end{equation*}
The execution result  $r_i$ is obtained by running the generated code $c_i$ to request the data from the backend servers of tools, denoted as $ r_i = \text{Execute}(c_i)$.
When the tool fails to execute, the $r_i$ indicates an error message as a failure signal.
When the tool executes successfully, the result $r_i$ contains the targeted information in response to the planning $t_i$.

\paragraph{Review Agent.}\label{sec:review}
Incorrect \textit{planning} and \textit{execution} are frequently observed during the tool-use workflow.
The \third \oa is employed as an expert, providing feedback to agent \ga and \ea for revision.
Specifically, if the planning generated by \ga is vague or selects a non-existing tool, the agent \oa generates verbal feedback to instruct the \ga to reformulate planning.
It can be formulated as:
\begin{equation}
\begin{aligned}
    f_{R\rightarrow G} &= M_R(x, \mathcal{S}, t_i)
\end{aligned}
\label{eq:frg}
\end{equation}
Similarly, if \ea hallucinates generating a wrong program to execute the tool, the agent \oa reviews execution results (or errors) and re-checks the tool documentation, providing instructions for calibration:
\begin{equation}
\begin{aligned}
    f_{R\rightarrow E} &= M_R \left(x, d, c_i, r_i\right)
\end{aligned}
\label{eq:fre}
\end{equation}
We denote the maximum turns of interaction between agent \oa and agent \ga (or \ea) is denoted as $\alpha$ (or $\beta$).
Their communication protocol and action flow are explained in \S~\ref{sec:protocol}.

\subsection{Agent communication protocols}\label{sec:protocol}

We propose two agent communication protocols, including \textit{automatic} and \textit{adaptive} interaction. 

\paragraph{Automatic interaction.}\label{sec:auto}
As illustrated in Figure~\ref{fig:method}, our automatic interaction alternates between  \textit{planning-review} and \textit{execution-review} phases.
For the $i$th step, it starts with the interaction between the agent \ga and \oa until a correct planning $t_i$ is determined or up to the maximum turns $\alpha$. 
Formally, it can be formulated as: 
\begin{equation}
\begin{aligned}
    t_i^j & = M_G (x, \mathcal{S}, \mathcal{H}_i,\underbrace{\{t_i^{<j}, f_{R\rightarrow G}^{<j}\}}_{\text{planning calibration}})
\end{aligned}
\label{eq:gr}
\end{equation}
Here, $j$ indicates $j$th interaction of two agents.
Following the planning $t$, the agent \ea generates executable programs to execute the selected tool and calibrates the incorrect result $r$ with the feedback of agent \oa for up to $\beta$ turns.
This process can be formulated as:
\begin{equation}
\begin{aligned}
 c_i^j  & = \mathcal{M}_E(t_i, d, \underbrace{\{c_i^{<j}, f_{R\rightarrow E}^{<j}\}}_{\text{execution calibration}})
\end{aligned}
\label{eq:er}
\end{equation}
The calibrated result is then incorporated into the context of \ga for the next planning generation.

\paragraph{Adaptive interaction.}

In our adaptive interaction strategy, the agent flow primarily alternates from (1) generating tool-use planning by agent \ga and (2) generating execution code by agent \ea, in a step-by-step manner.
The \third \oa is adaptively triggered to provide feedback only when the generated code fails to execute correctly.
Specifically, a runtime error can be caused by either unfeasible planning or coding faulty.
Thus, the agent \oa first reviews the generated planning and code, routines the errors to agent \ga or \ea accordingly, and provides feedback for revision.

\section{Specialization by Agent Distillation}\label{sec:calibration}

Our initial experiment shows that powerful LLMs such as GPT-4, achieve promising results when equipped with our framework.
However, these model are often considered black boxes~\citep{webcpm,gao2023confucius} with potential privacy issues.
Thus, we aim to adapt our framework to open-source models.
We propose \textit{specialized action distillation} (\distill), which distills the task-solving trajectory of powerful commercial LLMs into different open-source LLM agents tailored to specific functionalities in \ours.

\subsection{Synthesize the Training Dataset}

Our distillation method collects the task-solving trajectory of specialized agents simulated by GPT-4, in \ours (\S~\ref{sec:framework}).
To achieve this, we first sample tasks from ToolBench~\cite{toolllm}, which contains nearly 200k practical tasks across 3,451 tools.
We select 2,919 tasks using various heuristic strategies (see Appendix~\ref{sec:app:filter} for more details).
Each task $x$ is paired with a list of relevant tools.
Since we find that some tasks in ToolBench are very similar to each other, we cluster them based on the semantic similarities between task descriptions and retain one instance for each cluster.
Next, we supplement each of these selected tasks with a detailed solution.
Specifically, we separately implement our grounding, execution, and review agent with GPT-4, and coordinate them using the proposed automatic communication protocol (\S~\ref{sec:auto}) to generate solutions.
Finally, we synthesize a dataset with 500 diverse examples. 
Each example contains a task $x$, a candidate toolset $\mathcal{S}$, and the task-solving trajectory of three agents.
The statistics of our synthetic dataset are provided in Table~\ref{tab:dataset}.

\begin{table}[t]
\centering
\begin{adjustbox}{width=\columnwidth,center}
\begin{tabular}{p{8cm} r}
\toprule
\textbf{Statistic}
\\
\midrule
\# The data scale &  500
\\
\# The average tokens of input task  & 52.48
\\
\# The average number of candidate tools   &  20
\\
\# The average number of ground truth tools per task& 3.39
\\
\# The average turns of planning-review interaction& 4.62 
\\
\# The average turns of execution-review interaction& 5.21
\\
\bottomrule
\end{tabular}
\end{adjustbox}
\caption{The statistics of our synthetic dataset in our \textit{specialized action distillation} method.}
\label{tab:dataset}
\end{table}

\subsection{Agent Training}\label{sec:tuning}

Due to the large number of parameters of the LLM, we employ a parameter-efficient tuning technique (\ie LoRa~\citep{hu2021lora}) to train each specialized agent separately.
The objective is to optimize the delta parameters $\Delta\theta$ of the LLM $\theta$ to minimize the loss function.

We reorganize the dataset according to the agents' functionality (\S~\ref{sec:framework}), thereby distilling specific abilities into different student models.
Formally, given a task $x$, in the $i$th step, the $\mathcal{H}_i$ contains historical planning and execution results.
We train the agent \ga to generate the $i$th tool-use planning $t_i$ on the condition of $H_i$ and revise its incorrect planning following the review from agent \oa (\daulg{Eq.~\ref{eq:gr}}).
We train the agent \ea to generate programs $c$ for tool execution following the generated planning $t$ and feedback of agent \oa (\daulg{Eq.~\ref{eq:er}}).
Similarly, the agent \oa are trained to provide feedback as \daulg{Eq.~\ref{eq:frg} and Eq.~\ref{eq:fre}}.
We apply the standard language modeling loss for the optimization.
More details and formulations can be found in Appendix~\ref{sec:app:distillation}.

\section{Experimental Setup}\label{sec:experiment}

\begin{table*}[t!]
\centering
\renewcommand\arraystretch{1.0}
\scalebox{0.85}{
\begin{tabular}{l cc cc cc }
\toprule

{\multirow{2}{*}{\textbf{Method}}} 
& \multicolumn{2}{c}{\textbf{RestBench-TMDB}} 
& \multicolumn{2}{c}{\textbf{RestBench-Spotify}} 
& \multicolumn{2}{c}{\textbf{ToolBench}} \\
\cmidrule(lr){2-3} \cmidrule(lr){4-5} \cmidrule(lr){6-7} 
& \textbf{Success Rate} & \textbf{Path\%} 
& \textbf{Success Rate} & \textbf{Path\%}
& \textbf{Success Rate} & \textbf{Path\%} \\
\Xhline{1px}
\rowcolor{Gainsboro} \multicolumn{7}{l}{\textit{gpt-3.5-turbo}}
\\

{\small \faUser} ReAct~\citep{react}
& 40.00 & 71.19
&  51.28 & 60.35
& 39.39 & 65.04
 \\

{\small \faUser} Chameleon~\citep{chameleon}
& 63.00 & 66.10
& 56.20 & 64.55  
& 37.44 & 67.55  
\\

 {\small \faUser} CodeAct~\citep{wang2024executable}
& 63.00 &  80.91
& 54.30  & 76.64 
& -- &  --
\\

{\small \faUser} ToolLLM~\cite[DFSDT,][]{toolllm}
& 68.00 &  76.77
& 61.40 &  74.77
& 66.39 & \textbf{86.43}
\\

{\small \faUsers} Reflexion~\citep{reflexion} 
 & 53.00  & 55.00
 &   49.10 &   50.90
&   -- &  --
\\

{\small \faUsers} $\alpha$-UMi~\citep{shen2024small}
& 62.00  & 70.23 
& 66.74  & 70.27
& 67.55   & 78.37 
\\
{\small \faUsers} RestGPT~\citep{restgpt}
& 65.00  &  69.21   
&  67.10  & 70.75  
&  63.88  & 77.40  
\\

{\small \faUsers}  \textbf{\ours} \textit{w/ Ada}
& 78.00 & 79.57
& 69.43  & 77.54
& 69.84   &  81.58
\\

{\small \faUsers} \textbf{\ours} \textit{w/ Auto}
& \textbf{79.00} &  \textbf{81.97}
& \textbf{71.21} &  \textbf{79.17}
& \textbf{72.15}  & 83.33

\\
\midrule

{\small \faUser} ReAct@N $\rightarrow \text{N} =2 $
& 54.00  & 67.90
& 56.71  &  59.47
& 41.41 &63.67
 \\
 
{\small \faUser} ReAct@N $\rightarrow \text{N} =3 $
& 62.00 &  65.40
&  58.13 & 63.26
& 42.67 & 66.12   
\\

{\small \faUser} ToolLLM@N $\rightarrow \text{N} =2 $
& 70.00 &  76.54
& 63.16 &  75.27
&  68.37 & 86.43
\\

{\small \faUser} ToolLLM@N $\rightarrow \text{N} =3 $
& 71.00 &  78.11
&  63.16 &  76.30
& 68.77  & 87.54
\\

\bottomrule
\end{tabular}
}
\caption{
\textbf{The results on three datasets.}
The metrics Success\% and Path\% indicate the Success Rate and Correct Path Rate, respectively.
The icon {\small \faUser} denotes the single-agent method  and {\small \faUsers} symbolizes multi-agent architecture.
}
\label{tab:main}
\end{table*}

\subsection{Datasets and Evaluation Metrics}
\textbf{Datasets.} We conduct experiments on two well established benchmarks, \ie RestBench~\citep{restgpt} and Toolbench~\citep{toolllm}.
The RestBench consists of two subsets, including: (1) TMDB, a high-quality human annotated dataset consisting of 54 movie-related tools;
and (2) Spotify, a dataset with 40 music-related tools.
The Toolbench contains various practical tasks across diverse scenarios.
We provide more details for these datasets in Appendix~\ref{sec:app:dataset}.

\noindent\textbf{Evaluation metrics.}
Following~\citet{gpt4tools,gao2023confucius}, we use two evaluation metrics: 
(1) Success Rate (\textbf{Success\%}) measuring the proportion of successful query completions, and 
(2) Correct Path Rate (\textbf{Path\%}) calculating the F1 score between the generated tool sequence and ground-truth tool sequence.
We also conduct a human evaluation, in which three well-educated volunteers are invited to evaluate 30 randomly sampled cases with a three-scale rating in two aspects:
(1) Executability (Exec): whether multiple tools are invoked in a correct logical order; 
and (2) Utility: whether the execution results of tools can be used to generate an answer.

\subsection{Baselines}
We compare our method with agent-based tool learning methods, including:
(1) \textit{Chameleon}~\citep{chameleon}, 
an LLM-based agent that directly generates multi-step plans for tool use and then sequentially executes the plan; 
(2) \textit{ReAct}~\cite{react}, which prompts LLM to generate the chain-of-thought and actions in an interleaved manner.;
(3) \textit{CodeAct}~\cite{wang2024executable}, which allows the LLM to generate executable code snippets as actions to use tools;
(4) \textit{ToolLLM}~\cite[DFSDT,][]{toolllm}, which enhances LLMs with the Depth First Search-based Decision Tree (DFSDT) to select tools to solve a task.
For further comparison, 
Since our \ours coordinates three specialized agents, we also establish two baselines, \ie ReAct@N and ToolLLM@N, which are up to N times runs of their vanilla method (ReAct or ToolLLM) until an input task is completed.

We also consider baselines with multi-agent architecture, including (1) \textit{RestGPT}~\citep{restgpt}: which consists of a planning module, a tool selector, an executor, and a response parsing module;
(2) \textit{Reflexion}~\cite{reflexion}, which  employs an LLM for task execution and uses another LLM to verbally reflect on task feedback signals;
and (3) \textit{$\alpha$-UMi}~\cite{shen2024small}, which consists of a planner, an executor, and an answer generator.

\begin{table}[t!]
\centering
\setlength\tabcolsep{2pt}
\begin{adjustbox}{width=\columnwidth,center}
\renewcommand\arraystretch{1.0}
\begin{tabular}{ l cc cc }
\toprule
{\multirow{2}{*}{\textbf{Method}}} 
& \multicolumn{2}{c}{\textbf{TMDB}} 
& \multicolumn{2}{c}{\textbf{Spotify}} \\
\cmidrule(lr){2-3} \cmidrule(lr){4-5} 
& \textbf{Success\%} & \textbf{Path\%} 
& \textbf{Success\%} & \textbf{Path\%} \\
\Xhline{1px}

\rowcolor{Gainsboro} \multicolumn{5}{l}{{\small \faUsers} \ours (\textit{Mixtral-8x7B})} \\

~~\textit{w/ Auto} (Distilled)
&53.00& 79.32
& 36.09 & 73.92
\\

~~\textit{w/ Auto} (Vanilla)
&49.00& 76.22
& 34.21 & 68.14
\\

~~\textit{w/ Ada} (Distilled)
& 51.00& 78.74
&  35.47 & 69.86
\\

~~\textit{w/ Ada} (Vanilla)
& 47.00 & 74.05
&  33.33 & 66.41
\\

\hline
\rowcolor{Gainsboro} \multicolumn{5}{l}{Baselines (\textit{Mixtral-8x7B})} \\

{\small \faUser}ReAct
& 26.00 & 61.21
& 21.35 & 47.21
 \\

{\small \faUser}ReAct@3
& 33.00 & 63.27
& 26.93 & 50.31
\\

 {\small \faUser}ToolLLM
& 37.00 & 64.32
&  28.07 &   52.31
\\

 {\small \faUser}ToolLLM@3
& 45.00  & 74.40
&  31.58 &   57.68
\\

 {\small \faUsers} RestGPT
&  34.00 &  72.20
& 31.58 & 67.82
\\

\bottomrule
\end{tabular}
\end{adjustbox}
\caption{We employ the Mixtral-8x7B as the backbone LLM of for our method and baselines.
The \textit{Vanilla} and \textit{Distilled} indicate enable our framework by prompting and our action distillation, respectively.
}
\label{tab:other-model}
\end{table}

\subsection{Implementation Details}
We use \textit{gpt-3.5-turbo}\footnote{\url{https://openai.com/chatgpt}} from OpenAI as the LLM backbone for each agent in our method and all baselines. 
We instruct the three agents to perform specific actions with different system prompts.
The decoding temperature is set to 0 for the most deterministic generation. 
We also repeat the experiment with an open-source model Mistral-8x7B\footnote{\href{https://huggingface.co/mistralai/Mixtral-8x7B-Instruct-v0.1}{https://huggingface.co/mistralai}} for further comparison.
In our agent communication (\S~\ref{sec:protocol}), we set the maximum iteration of interactions $\alpha = 3$ and $\beta = 3$, respectively.
For each sample in the test set,  we provide all the baselines with the same candidate toolset for a fair comparison, which contains the required tools and ten randomly sampled tools.

Our action distillation separately trains three Mistral-8x7B using the corresponding optimization objectives in \S~\ref{sec:tuning} with the learning rate of $5\times 10^{-5}$.
The training of our model can be done within 4 hours with 3 NVIDIA A800-PCIE-80GB GPUs using LoRA~\cite{hu2021lora}.

\section{Results and Analysis}\label{sec:result}

\subsection{Experimental Results}

\paragraph{Overall performance.}
\label{sec:over-performance}
Table~\ref{tab:main} demonstrates the experimental performances of all methods.
We find that our proposed \ours outperforms all the baselines in three datasets in terms of all metrics.
A reason here is that our cooperative framework design enables each agent to perform specialized actions instead of grasping all required capabilities, thereby reducing the workload encountered by a single agent.
The significant improvement over ReAct@N and ToolLLM@N baselines can further validate the effectiveness of our framework.
Compared with baselines with multi-agent architecture like RestGPT, \ours achieves about 12\% higher Success Rate.
The potential reason for our improvement is that the proposed two communication protocols enable the dynamic interaction of agents, which is more flexible to handle exception errors.

\paragraph{Performance with the open-source LLM.}
We further evaluate our \ours by swapping the backbone LLM with  Mistral-8x7B and repeating the experiment under the same conditions.
As shown in Table~\ref{tab:other-model}, we implement our framework in two ways with Mistral-8x7B:
(1) directly prompting (\textit{w/ Auto} and \textit{w/ Ada});
(2) tuning with our proposed action distillation (\textit{w/ Auto}$^\dagger$ and \textit{w/ Ada}$^\dagger$).
We observe that directly prompting Mistral-8x7B with \ours yields better performance than baselines.
The action distillation further improves overall performance substantially, such as pushing the Success Rate from 47.00 to 51.00 in the TMDB dataset.
These results further prove the effectiveness of our cooperative framework.

\subsection{Human Evaluation}\label{sec:human-evaluation}
Table~\ref{tab:human} shows the results of the human evaluation.
We find that \ours achieves the best results in the Executability aspect with 0.08\textasciitilde 0.12 improvement.
These results further validate the necessity of agent specialization and cooperation.
The overall Kappa statistics for Executability and Utility are $0.75$ and $0.71$, illustrating substantial agreement~\cite{landis1977measurement} among the annotators.

\begin{table}[!t]
\centering
\begin{adjustbox}{width=\columnwidth,center}
\setlength\tabcolsep{4pt}
\begin{tabular}{@{}p{2.6 cm} ccc c@{}}

\toprule

{\multirow{2}{*}{\textbf{Method}}} 
& \multicolumn{2}{c}{\textbf{TMDB}} 
& \multicolumn{2}{c}{\textbf{Spotify}} \\
\cmidrule(lr){2-3} \cmidrule(lr){4-5} 
& \textbf{Success\%} & \textbf{Path\%} 
& \textbf{Success\%} & \textbf{Path\%} \\

\midrule
\textit{Ours w/ Auto}
& 79.00 & 81.97 
& 71.43 & 77.54 
\\

\cdashline{1-5}[6pt/6pt]
\specialrule{0em}{1pt}{1pt}

\textit{w/o} \oa$\rightarrow$\ga
& 77.00$\downarrow_{2.0}$  &  78.10$\downarrow_{3.9}$ 
& 68.42$\downarrow_{3.0}$  &  75.33$\downarrow_{2.2}$ 
\\

\textit{w/o} \oa$\rightarrow$\ea
& 75.00$\downarrow_{4.0}$  & 74.23$\downarrow_{7.7}$
& 64.91$\downarrow_{6.5}$ & 72.41$\downarrow_{5.1}$
\\

\textit{w/ static coop.}
& 75.00$\downarrow_{4.00}$  & 75.74$\downarrow_{6.2}$
& 67.12$\downarrow_{4.3}$ &  75.07 $\downarrow_{2.5}$

\\

\bottomrule
\end{tabular}
\end{adjustbox}

\caption{The ablation study on two datasets with \textit{gpt-3.5-turbo} as backbone. See \S~\ref{sec:ablation} for details}
\label{tab:ablation}
\end{table}

\subsection{Ablation Study}\label{sec:ablation}
To better understand the impact of different components of our method, we make the following modifications to the architecture and measure the effect.

\noindent\textbf{- w/o} \oa $\rightarrow$ \ga.
We remove the interaction between agent \oa and \ga in our framework.
As shown in Table~\ref{tab:ablation}, the Success Rate has a average 2.50 decline, while the Correct Path Rate has a 3.05 average decline on two datasets.
This results validate the necessity of feedback of \oa which can instruct the \ga to revise incorrect planning.

\noindent\textbf{- w/o} \oa $\rightarrow$ \ea.
We remove the interaction between agent \oa and \ea in our framework when programming to execute tools.
As shown in Table~\ref{tab:ablation}, the Success Rate suffers from obvious decrease in both two datasets.
These results indicate that the agent \oa can review the generated programs of agent \ea and provide useful instruction for calibrating errors.

\noindent\textbf{- w/ static cooperation}.
We implement the \oa with a code compiler, which is triggered to provide static feedback only when runtime errors are raised during executing tools by agent \ea.
This allows us to compare our framework with a static algorithm for agent cooperation. 
Table~\ref{tab:ablation} present the results, where we observe a 4.12 average decrease in the Success Rate, \eg dropping from 79.00 to 75.00 on the TMDB dataset.
The same trend is also observed in the Correct Path Rate, \eg a 2.5 decrease on the Spotify dataset.
These results indicate the superiority of our dynamic agent cooperation framework.

\begin{table}[t!]
\centering
\begin{adjustbox}{width=\columnwidth,center}
\renewcommand\arraystretch{1.0}
\begin{tabular}{ l cc cc }
\toprule
{\multirow{2}{*}{\textbf{Method}}} 
& \multicolumn{2}{c}{\textbf{TMDB}} 
& \multicolumn{2}{c}{\textbf{Spotify}} \\
\cmidrule(lr){2-3} \cmidrule(lr){4-5} 
& \textbf{Exec} & \textbf{Utility} 
& \textbf{Exec} & \textbf{Utility} \\
\Xhline{1px}

\rowcolor{Gainsboro} \multicolumn{5}{l}{\textit{gpt-3.5-turbo}} \\

 {\small \faUser} ReAct
&  1.89 &  1.93
& 1.77  & 2.10
\\

{\small \faUser} ToolLLM
& 2.26  & 1.87
& 2.26  & 2.30
\\

 {\small \faUsers} RestGPT 
& 2.35 & 2.45
& 2.30  &2.40
\\

{\small \faUsers} Ours \textit{w/ Auto}
& 2.47  & 2.56
& 2.43  & 2.50
\\
{\small \faUsers} Ours \textit{w/ Ada}
& 2.43 &2.50
& 2.38  & 2.45
\\

\bottomrule
\end{tabular}
\end{adjustbox}
\caption{Human evaluation on Executability (\textbf{Exec}) and Correct Rate of Parsing (\textbf{Parsing}).  }
\label{tab:human}
\end{table}

\subsection{Case Study}
We conduct the case studies and find that our cooperative agent framework is more effective at executing various tools and handle exceptional errors in solving tasks.
We also provide  examples to explain the detailed process of agent cooperation.
The details can be found in Appendix~\ref{sec:app:case}.

\section{Discussion}\label{sec:discussion}

\begin{figure}[t]
        \centering
\includegraphics[width=0.475\textwidth]{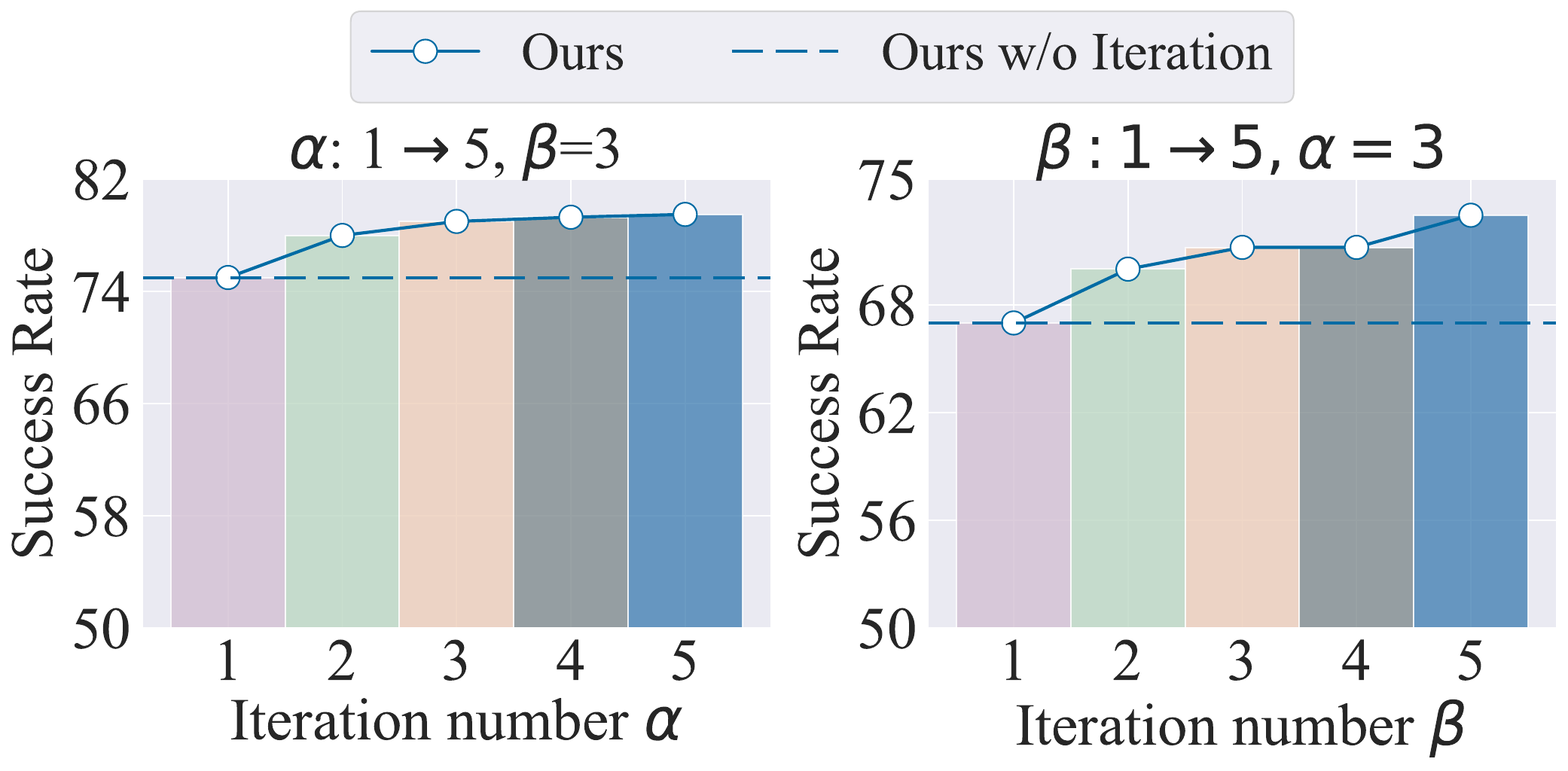}
        \caption{The qualitative analysis for the maximum interaction turns  $\alpha$ and $\beta$ in our agent communication protocols (Section~\ref{sec:protocol}) on the TMDB dataset.}
 \label{fig:iteration}
\end{figure}

\noindent \paragraph{Qualitative analysis for the maximum number of interactions.} 
In our \textit{automatic agent interaction}, agents \ga and \ea revise their actions following the feedback of agent \oa for up to $\alpha$ and $\beta$ turns, respectively.
To further explore the impact of the interaction times on overall performance, we conduct a quantitative and qualitative analysis by varying $\alpha$ and $\beta$ from 1 to 5.
Then we evaluate our framework using the RestBench-TMDB dataset with the same settings as in Table~\ref{tab:main}.
As illustrated in Figure~\ref{fig:iteration}, we find an increasing Success Rate when the maximum iteration turns shifts from 1 to 3.
In addition, a relatively stable trend is observed when the $\alpha$ and  $\beta$ keep increasing (from 3 to 5), which indicates the agents can correct most errors within 3 turns.
We also look at the poorly performing cases where we find that since the planning from agent \ga is typically open-ended, the \oa struggles to detect all the incorrect planning.
For example, the planning may be plausible and clear but lacks the required arguments to execute tools, thus resulting in a failure of \ea in subsequent steps.

\begin{figure}[t]
        \centering
\includegraphics[width=0.5\textwidth]{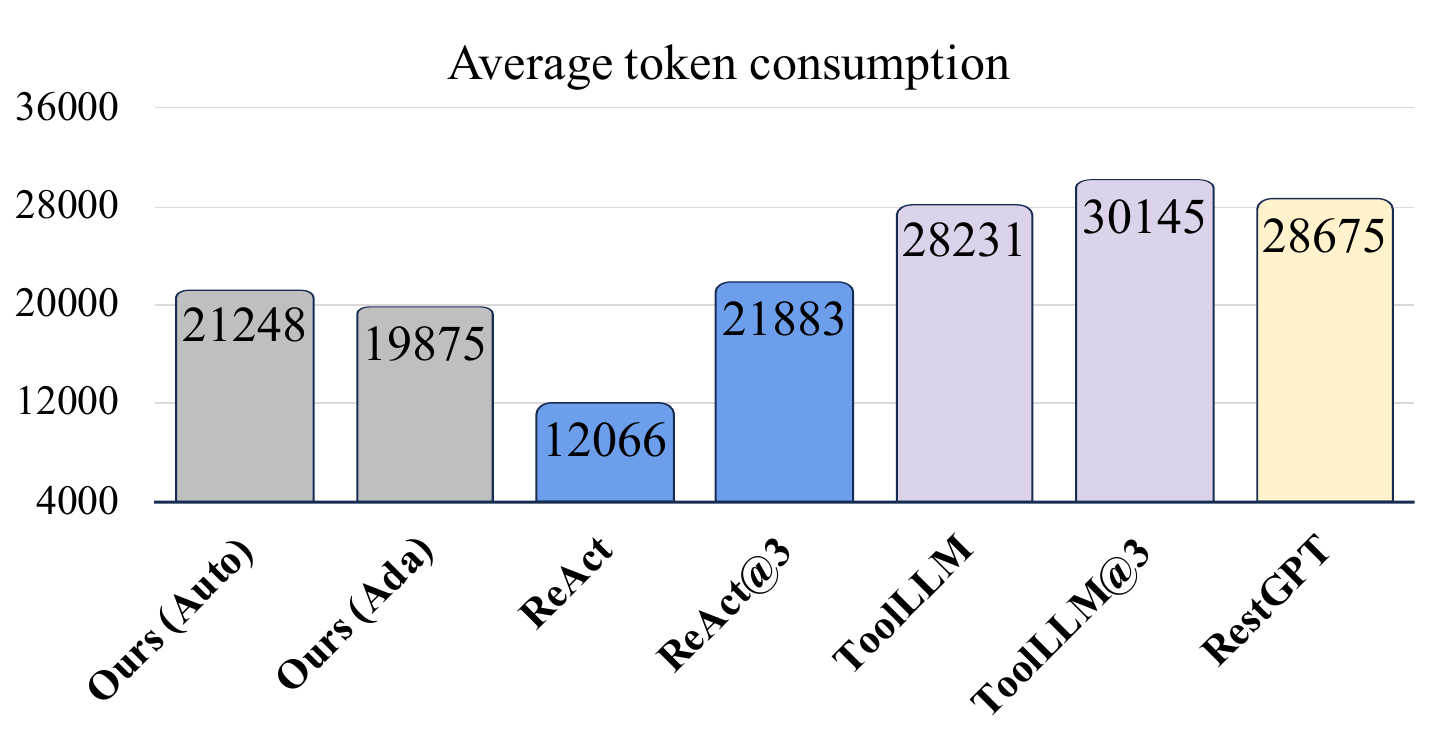}
        \caption{
The efficiency analysis for different methods, where we count the average consumed tokens.
}
 \label{fig:token}
\end{figure}

\noindent  \paragraph{Qualitative analysis for the efficiency of inference.}
Due to the intensive inference cost of LLMs-based agents, we further explore the efficiency of our  \ours.
To explain more intuitively, 
we compare the token consumption for the \ours and baselines using the RestBench-TMDB dataset with the same settings as in Table~\ref{tab:main}.
As illustrated in Figure~\ref{fig:token}, we find that 
although our framework achieves better performance, we spend fewer tokens compared with strong baselines such as RestGPT and ToolLLM@3.
The reason is that the cooperative framework \ours enables each agent to perform specific tasks more efficiently, reducing the length exploration trajectory by the single agent. 

\noindent \paragraph{The quality of generated review.}
We further analyze the quality of reviews given by \third \oa.
Specifically, we randomly sample 50 task-solving trajectories in Table~\ref{tab:main} (w/ Auto) manually analyze the review of \third.
For most tasks, we find that the agent \oa can assist agent \ea to revise its generated code or provides useful reviews for the planning generated by agent \ga, such as \texttt{only select tools from given list}.  
In addition, we find that in less than 5\% of tasks, the agent \oa hallucinates giving an incorrect review, indicating its reliability.

\noindent \paragraph{Runtime consistency.}
Considering the non-deterministic nature of LLM generation, we  analyze the consistency of our framework.
We repeat our method multiple times with the same settings as in Table~\ref{tab:main}.
The statistical significance of differences observed between the performance of two runs is tested using a two-tailed paired t-test.
We find no significant difference between the results of two randomly conducted experiments (significance level $\alpha$ = 0.05).

\section{Conclusions}\label{sec:conclusion}

We present a cooperative and interactive agents framework (\ours) for tool learning, which diverges from previous work by allowing the cooperation of agents to solve complex tasks.
The \ours first modularizes the overall workflow with three specialized agents for tool planning, tool execution, and action calibration, respectively.
Then, two communication protocols are introduced to enable the dynamic cooperation of these agents.
To generalize our framework to open-source models, we propose specialized action distillation, enhancing the models' capability to perform specific actions.
Extensive experiments conducted on three datasets demonstrate the superiority of our \ours, \eg pushing the success rate to 77.00 with 13.2\% point improvement.
Our future work includes:
(1) extending our method to agents empowered by multi-modal foundation models, incorporating image and sound;
(2) coordinating the cooperation between strong and weak agents.

\section*{Limitations}\label{sec:lmitation}
The main limitation is that our LLM-based agent is limited when perceiving multi-modal tasks. When executing the tools, we represent the image and speech input with \code{url}, following previous works.
In the future, we plan to extend our method to agents empowered by multi-modal foundation models.

\section*{Ethics Statement}
The paper proposes a cooperative agent framework, synergizing specialized agents to solve complex tasks.
The modularized design enables the agents to utilize feedback from the tool environment to calibrate themselves adaptively.
In addition to the use of state-of-the-art commercial LLMs, we have experimented with an open-source LLM, for reproducibility reasons and to allow the use of our method in lower-resource contexts. 
All the tools used in our experiment are provided by open-source platforms, including TMDB, Spotify, and Rapid API, thus ensuring a high level of transparency and reproducibility.

We have made every effort to ensure that our research does not harm individuals or groups, nor does it involve any form of deception or potential misuse of information. 

\bibliography{custom}

\begin{thebibliography}{46}
\expandafter\ifx\csname natexlab\endcsname\relax\def\natexlab#1{#1}\fi

\bibitem[{Bran et~al.(2023)Bran, Cox, White, and Schwaller}]{chemcrow}
Andres~M Bran, Sam Cox, Andrew~D White, and Philippe Schwaller. 2023.
\newblock Chemcrow: Augmenting large-language models with chemistry tools.
\newblock \emph{arXiv preprint arXiv:2304.05376}.

\bibitem[{Chang et~al.(2023)Chang, Wang, Wang, Wu, Yang, Zhu, Chen, Yi, Wang, Wang et~al.}]{chang2023survey}
Yupeng Chang, Xu~Wang, Jindong Wang, Yuan Wu, Linyi Yang, Kaijie Zhu, Hao Chen, Xiaoyuan Yi, Cunxiang Wang, Yidong Wang, et~al. 2023.
\newblock A survey on evaluation of large language models.
\newblock \emph{ACM Transactions on Intelligent Systems and Technology}.

\bibitem[{Cohen et~al.(2023)Cohen, Hamri, Geva, and Globerson}]{cohen2023lm}
Roi Cohen, May Hamri, Mor Geva, and Amir Globerson. 2023.
\newblock Lm vs lm: Detecting factual errors via cross examination.
\newblock \emph{arXiv preprint arXiv:2305.13281}.

\bibitem[{Du et~al.(2023)Du, Li, Torralba, Tenenbaum, and Mordatch}]{du2023improving}
Yilun Du, Shuang Li, Antonio Torralba, Joshua~B Tenenbaum, and Igor Mordatch. 2023.
\newblock Improving factuality and reasoning in language models through multiagent debate.
\newblock \emph{arXiv preprint arXiv:2305.14325}.

\bibitem[{Dziri et~al.(2023)Dziri, Lu, Sclar, Li, Jian, Lin, West, Bhagavatula, Bras, Hwang et~al.}]{dziri2023faith}
Nouha Dziri, Ximing Lu, Melanie Sclar, Xiang~Lorraine Li, Liwei Jian, Bill~Yuchen Lin, Peter West, Chandra Bhagavatula, Ronan~Le Bras, Jena~D Hwang, et~al. 2023.
\newblock Faith and fate: Limits of transformers on compositionality.
\newblock \emph{arXiv preprint arXiv:2305.18654}.

\bibitem[{Fu et~al.(2023)Fu, Peng, Khot, and Lapata}]{fu2023improving}
Yao Fu, Hao Peng, Tushar Khot, and Mirella Lapata. 2023.
\newblock Improving language model negotiation with self-play and in-context learning from ai feedback.
\newblock \emph{arXiv preprint arXiv:2305.10142}.

\bibitem[{Gao et~al.(2023{\natexlab{a}})Gao, Madaan, Zhou, Alon, Liu, Yang, Callan, and Neubig}]{pal}
Luyu Gao, Aman Madaan, Shuyan Zhou, Uri Alon, Pengfei Liu, Yiming Yang, Jamie Callan, and Graham Neubig. 2023{\natexlab{a}}.
\newblock {PAL}: Program-aided language models.
\newblock In \emph{PMLR}, pages 10764--10799.

\bibitem[{Gao et~al.(2023{\natexlab{b}})Gao, Shi, Zhu, Fang, Xin, Ren, Chen, Ma, and Ren}]{gao2023confucius}
Shen Gao, Zhengliang Shi, Minghang Zhu, Bowen Fang, Xin Xin, Pengjie Ren, Zhumin Chen, Jun Ma, and Zhaochun Ren. 2023{\natexlab{b}}.
\newblock \href {http://arxiv.org/abs/2308.14034} {Confucius: Iterative tool learning from introspection feedback by easy-to-difficult curriculum}.

\bibitem[{Hu et~al.(2021)Hu, Shen, Wallis, Allen-Zhu, Li, Wang, Wang, and Chen}]{hu2021lora}
Edward~J Hu, Yelong Shen, Phillip Wallis, Zeyuan Allen-Zhu, Yuanzhi Li, Shean Wang, Lu~Wang, and Weizhu Chen. 2021.
\newblock Lora: Low-rank adaptation of large language models.
\newblock \emph{arXiv preprint arXiv:2106.09685}.

\bibitem[{Kim et~al.(2023)Kim, Baldi, and McAleer}]{rci}
Geunwoo Kim, Pierre Baldi, and Stephen McAleer. 2023.
\newblock Language models can solve computer tasks.
\newblock \emph{ArXiv}, abs/2303.17491.

\bibitem[{Landis and Koch(1977)}]{landis1977measurement}
J~Richard Landis and Gary~G Koch. 1977.
\newblock The measurement of observer agreement for categorical data.
\newblock \emph{biometrics}, pages 159--174.

\bibitem[{Liang et~al.(2023)Liang, He, Jiao, Wang, Wang, Wang, Yang, Tu, and Shi}]{liang2023encouraging}
Tian Liang, Zhiwei He, Wenxiang Jiao, Xing Wang, Yan Wang, Rui Wang, Yujiu Yang, Zhaopeng Tu, and Shuming Shi. 2023.
\newblock Encouraging divergent thinking in large language models through multi-agent debate.
\newblock \emph{arXiv preprint arXiv:2305.19118}.

\bibitem[{Liu et~al.(2023)Liu, Zhang, Li, Liu, and Yang}]{liu2023dynamic}
Zijun Liu, Yanzhe Zhang, Peng Li, Yang Liu, and Diyi Yang. 2023.
\newblock Dynamic llm-agent network: An llm-agent collaboration framework with agent team optimization.
\newblock \emph{arXiv preprint arXiv:2310.02170}.

\bibitem[{Lu et~al.(2023)Lu, Peng, Cheng, Galley, Chang, Wu, Zhu, and Gao}]{chameleon}
Pan Lu, Baolin Peng, Hao Cheng, Michel Galley, Kai-Wei Chang, Ying~Nian Wu, Song-Chun Zhu, and Jianfeng Gao. 2023.
\newblock Chameleon: Plug-and-play compositional reasoning with large language models.
\newblock \emph{ArXiv}, abs/2304.09842.

\bibitem[{Mallen et~al.(2023)Mallen, Asai, Zhong, Das, Khashabi, and Hajishirzi}]{mallen2023not}
Alex Mallen, Akari Asai, Victor Zhong, Rajarshi Das, Daniel Khashabi, and Hannaneh Hajishirzi. 2023.
\newblock When not to trust language models: Investigating effectiveness of parametric and non-parametric memories.
\newblock In \emph{Proceedings of the 61st Annual Meeting of the Association for Computational Linguistics (Volume 1: Long Papers)}, pages 9802--9822.

\bibitem[{Mohtashami et~al.(2023)Mohtashami, Hartmann, Gooding, Zilka, Sharifi et~al.}]{mohtashami2023social}
Amirkeivan Mohtashami, Florian Hartmann, Sian Gooding, Lukas Zilka, Matt Sharifi, et~al. 2023.
\newblock Social learning: Towards collaborative learning with large language models.
\newblock \emph{arXiv preprint arXiv:2312.11441}.

\bibitem[{Nakano et~al.(2021)Nakano, Hilton, Balaji, Wu, Ouyang, Kim, Hesse, Jain, Kosaraju, Saunders, Jiang, Cobbe, Eloundou, Krueger, Button, Knight, Chess, and Schulman}]{webgpt}
Reiichiro Nakano, Jacob Hilton, S.~Arun Balaji, Jeff Wu, Long Ouyang, Christina Kim, Christopher Hesse, Shantanu Jain, Vineet Kosaraju, William Saunders, Xu~Jiang, Karl Cobbe, Tyna Eloundou, Gretchen Krueger, Kevin Button, Matthew Knight, Benjamin Chess, and John Schulman. 2021.
\newblock Webgpt: Browser-assisted question-answering with human feedback.
\newblock \emph{ArXiv}, abs/2112.09332.

\bibitem[{Paranjape et~al.(2023)Paranjape, Lundberg, Singh, Hajishirzi, Zettlemoyer, and Ribeiro}]{art}
Bhargavi Paranjape, Scott~M. Lundberg, Sameer Singh, Hanna Hajishirzi, Luke Zettlemoyer, and Marco~Tulio Ribeiro. 2023.
\newblock Art: Automatic multi-step reasoning and tool-use for large language models.
\newblock \emph{ArXiv}, abs/2303.09014.

\bibitem[{Patil et~al.(2023)Patil, Zhang, Wang, and Gonzalez}]{patil2023gorilla}
Shishir~G Patil, Tianjun Zhang, Xin Wang, and Joseph~E Gonzalez. 2023.
\newblock Gorilla: Large language model connected with massive apis.
\newblock \emph{arXiv preprint arXiv:2305.15334}.

\bibitem[{Prasad et~al.(2023)Prasad, Koller, Hartmann, Clark, Sabharwal, Bansal, and Khot}]{prasad2023adapt}
Archiki Prasad, Alexander Koller, Mareike Hartmann, Peter Clark, Ashish Sabharwal, Mohit Bansal, and Tushar Khot. 2023.
\newblock Adapt: As-needed decomposition and planning with language models.
\newblock \emph{arXiv preprint arXiv:2311.05772}.

\bibitem[{Qian et~al.(2023)Qian, Cong, Yang, Chen, Su, Xu, Liu, and Sun}]{qian2023communicative}
Chen Qian, Xin Cong, Cheng Yang, Weize Chen, Yusheng Su, Juyuan Xu, Zhiyuan Liu, and Maosong Sun. 2023.
\newblock Communicative agents for software development.
\newblock \emph{arXiv preprint arXiv:2307.07924}.

\bibitem[{Qiao et~al.(2024)Qiao, Zhang, Fang, Luo, Zhou, Jiang, Lv, and Chen}]{qiao2024autoact}
Shuofei Qiao, Ningyu Zhang, Runnan Fang, Yujie Luo, Wangchunshu Zhou, Yuchen~Eleanor Jiang, Chengfei Lv, and Huajun Chen. 2024.
\newblock Autoact: Automatic agent learning from scratch via self-planning.
\newblock \emph{arXiv preprint arXiv:2401.05268}.

\bibitem[{Qin et~al.(2023{\natexlab{a}})Qin, Cai, Jin, Yan, Liang, Zhu, Lin, Han, Ding, Wang, Xie, Qi, Liu, Sun, and Zhou}]{webcpm}
Yujia Qin, Zihan Cai, Dian Jin, Lan Yan, Shihao Liang, Kunlun Zhu, Yankai Lin, Xu~Han, Ning Ding, Huadong Wang, Ruobing Xie, Fanchao Qi, Zhiyuan Liu, Maosong Sun, and Jie Zhou. 2023{\natexlab{a}}.
\newblock {W}eb{CPM}: Interactive web search for {C}hinese long-form question answering.
\newblock In \emph{ACL}, pages 8968--8988.

\bibitem[{Qin et~al.(2023{\natexlab{b}})Qin, Hu, Lin, Chen, Ding, Cui, Zeng, Huang, Xiao, Han, Fung, Su, Wang, Qian, Tian, Zhu, Liang, Shen, Xu, Zhang, Ye, Li, Tang, Yi, Zhu, Dai, Yan, Cong, Lu, Zhao, Huang, Yan, Han, Sun, Li, Phang, Yang, Wu, Ji, Liu, and Sun}]{toolw}
Yujia Qin, Shengding Hu, Yankai Lin, Weize Chen, Ning Ding, Ganqu Cui, Zheni Zeng, Yufei Huang, Chaojun Xiao, Chi Han, Yi~Ren Fung, Yusheng Su, Huadong Wang, Cheng Qian, Runchu Tian, Kunlun Zhu, Shi Liang, Xingyu Shen, Bokai Xu, Zhen Zhang, Yining Ye, Bo~Li, Ziwei Tang, Jing Yi, Yu~Zhu, Zhenning Dai, Lan Yan, Xin Cong, Ya-Ting Lu, Weilin Zhao, Yuxiang Huang, Jun-Han Yan, Xu~Han, Xian Sun, Dahai Li, Jason Phang, Cheng Yang, Tongshuang Wu, Heng Ji, Zhiyuan Liu, and Maosong Sun. 2023{\natexlab{b}}.
\newblock Tool learning with foundation models.
\newblock \emph{ArXiv}, abs/2304.08354.

\bibitem[{Qin et~al.(2024)Qin, Liang, Ye, Zhu, Yan, Lu, Lin, Cong, Tang, Qian, Zhao, Tian, Xie, Zhou, Gerstein, Li, Liu, and Sun}]{toolllm}
Yujia Qin, Shi Liang, Yining Ye, Kunlun Zhu, Lan Yan, Ya-Ting Lu, Yankai Lin, Xin Cong, Xiangru Tang, Bill Qian, Sihan Zhao, Runchu Tian, Ruobing Xie, Jie Zhou, Marc~H. Gerstein, Dahai Li, Zhiyuan Liu, and Maosong Sun. 2024.
\newblock Toolllm: Facilitating large language models to master 16000+ real-world apis.
\newblock In \emph{ICLR}.

\bibitem[{Qu et~al.(2024)Qu, Dai, Wei, Cai, Wang, Yin, Xu, and Wen}]{qu2024tool}
Changle Qu, Sunhao Dai, Xiaochi Wei, Hengyi Cai, Shuaiqiang Wang, Dawei Yin, Jun Xu, and Ji-Rong Wen. 2024.
\newblock Tool learning with large language models: A survey.
\newblock \emph{arXiv preprint arXiv:2405.17935}.

\bibitem[{Schick et~al.(2023)Schick, Dwivedi-Yu, Dess{\`i}, Raileanu, Lomeli, Zettlemoyer, Cancedda, and Scialom}]{toolformer}
Timo Schick, Jane Dwivedi-Yu, Roberto Dess{\`i}, Roberta Raileanu, Maria Lomeli, Luke Zettlemoyer, Nicola Cancedda, and Thomas Scialom. 2023.
\newblock Toolformer: Language models can teach themselves to use tools.
\newblock \emph{ArXiv}, abs/2302.04761.

\bibitem[{Shen et~al.(2024)Shen, Li, Chen, Yan, Quan, Chen, Zhang, and Huang}]{shen2024small}
Weizhou Shen, Chenliang Li, Hongzhan Chen, Ming Yan, Xiaojun Quan, Hehong Chen, Ji~Zhang, and Fei Huang. 2024.
\newblock Small llms are weak tool learners: A multi-llm agent.
\newblock \emph{arXiv preprint arXiv:2401.07324}.

\bibitem[{Shen et~al.(2023)Shen, Song, Tan, Li, Lu, and Zhuang}]{huggingfacegpt}
Yongliang Shen, Kaitao Song, Xu~Tan, Dong~Sheng Li, Weiming Lu, and Yue~Ting Zhuang. 2023.
\newblock Hugginggpt: Solving ai tasks with chatgpt and its friends in huggingface.
\newblock \emph{ArXiv}, abs/2303.17580.

\bibitem[{Shi et~al.(2023)Shi, Chen, Misra, Scales, Dohan, Chi, Sch\"{a}rli, and Zhou}]{pmlr-v202-shi23a}
Freda Shi, Xinyun Chen, Kanishka Misra, Nathan Scales, David Dohan, Ed~H. Chi, Nathanael Sch\"{a}rli, and Denny Zhou. 2023.
\newblock Large language models can be easily distracted by irrelevant context.
\newblock In \emph{Proceedings of the 40th International Conference on Machine Learning}, Proceedings of Machine Learning Research. PMLR.

\bibitem[{Shi et~al.(2024)Shi, Gao, Chen, Feng, Yan, Shi, Yin, Chen, Verberne, and Ren}]{shi2024chain}
Zhengliang Shi, Shen Gao, Xiuyi Chen, Yue Feng, Lingyong Yan, Haibo Shi, Dawei Yin, Zhumin Chen, Suzan Verberne, and Zhaochun Ren. 2024.
\newblock Chain of tools: Large language model is an automatic multi-tool learner.
\newblock \emph{arXiv preprint arXiv:2405.16533}.

\bibitem[{Shinn et~al.(2023)Shinn, Cassano, Labash, Gopinath, Narasimhan, and Yao}]{reflexion}
Noah Shinn, Federico Cassano, Beck Labash, Ashwin Gopinath, Karthik Narasimhan, and Shunyu Yao. 2023.
\newblock Reflexion: Language agents with verbal reinforcement learning.(2023).
\newblock \emph{arXiv preprint cs.AI/2303.11366}.

\bibitem[{Song et~al.(2023)Song, Xiong, Zhu, Li, Wang, Tian, and Li}]{restgpt}
Yifan Song, Weimin Xiong, Dawei Zhu, Chengzu Li, Ke~Wang, Ye~Tian, and Sujian Li. 2023.
\newblock Restgpt: Connecting large language models with real-world applications via restful apis.
\newblock \emph{ArXiv}, abs/2306.06624.

\bibitem[{Sun et~al.(2023)Sun, Yin, Li, Wu, Qiu, and Kong}]{sun2023corex}
Qiushi Sun, Zhangyue Yin, Xiang Li, Zhiyong Wu, Xipeng Qiu, and Lingpeng Kong. 2023.
\newblock Corex: Pushing the boundaries of complex reasoning through multi-model collaboration.
\newblock \emph{arXiv preprint arXiv:2310.00280}.

\bibitem[{Talebirad and Nadiri(2023)}]{talebirad2023multi}
Yashar Talebirad and Amirhossein Nadiri. 2023.
\newblock Multi-agent collaboration: Harnessing the power of intelligent llm agents.
\newblock \emph{arXiv preprint arXiv:2306.03314}.

\bibitem[{Wang et~al.(2023{\natexlab{a}})Wang, Ren, Yang, Liang, Bai, Zhang, Yan, and Li}]{wang2023mac}
Bing Wang, Changyu Ren, Jian Yang, Xinnian Liang, Jiaqi Bai, Qian-Wen Zhang, Zhao Yan, and Zhoujun Li. 2023{\natexlab{a}}.
\newblock Mac-sql: Multi-agent collaboration for text-to-sql.
\newblock \emph{arXiv preprint arXiv:2312.11242}.

\bibitem[{Wang et~al.(2024{\natexlab{a}})Wang, Chen, Yuan, Zhang, Li, Peng, and Ji}]{wang2024executable}
Xingyao Wang, Yangyi Chen, Lifan Yuan, Yizhe Zhang, Yunzhu Li, Hao Peng, and Heng Ji. 2024{\natexlab{a}}.
\newblock \href {http://arxiv.org/abs/2402.01030} {Executable code actions elicit better llm agents}.
\newblock In \emph{ICML}.

\bibitem[{Wang et~al.(2023{\natexlab{b}})Wang, Wang, Liu, Chen, Yuan, Peng, and Ji}]{wang2023mint}
Xingyao Wang, Zihan Wang, Jiateng Liu, Yangyi Chen, Lifan Yuan, Hao Peng, and Heng Ji. 2023{\natexlab{b}}.
\newblock Mint: Evaluating llms in multi-turn interaction with tools and language feedback.
\newblock \emph{arXiv preprint arXiv:2309.10691}.

\bibitem[{Wang et~al.(2023{\natexlab{c}})Wang, Kordi, Mishra, Liu, Smith, Khashabi, and Hajishirzi}]{self-instruction}
Yizhong Wang, Yeganeh Kordi, Swaroop Mishra, Alisa Liu, Noah~A. Smith, Daniel Khashabi, and Hannaneh Hajishirzi. 2023{\natexlab{c}}.
\newblock Self-instruct: Aligning language models with self-generated instructions.
\newblock In \emph{ACL}.

\bibitem[{Wang et~al.(2024{\natexlab{b}})Wang, Cheng, Zhu, Fried, and Neubig}]{wang2024tools}
Zhiruo Wang, Zhoujun Cheng, Hao Zhu, Daniel Fried, and Graham Neubig. 2024{\natexlab{b}}.
\newblock What are tools anyway? a survey from the language model perspective.
\newblock \emph{arXiv preprint arXiv:2403.15452}.

\bibitem[{Yang et~al.(2023{\natexlab{a}})Yang, Song, Li, Zhao, Ge, Li, and Shan}]{gpt4tools}
Rui Yang, Lin Song, Yanwei Li, Sijie Zhao, Yixiao Ge, Xiu Li, and Ying Shan. 2023{\natexlab{a}}.
\newblock Gpt4tools: Teaching large language model to use tools via self-instruction.
\newblock \emph{ArXiv}, abs/2305.18752.

\bibitem[{Yang et~al.(2023{\natexlab{b}})Yang, Li, Wang, Lin, Azarnasab, Ahmed, Liu, Liu, Zeng, and Wang}]{yang2023mm}
Zhengyuan Yang, Linjie Li, Jianfeng Wang, Kevin Lin, Ehsan Azarnasab, Faisal Ahmed, Zicheng Liu, Ce~Liu, Michael Zeng, and Lijuan Wang. 2023{\natexlab{b}}.
\newblock Mm-react: Prompting chatgpt for multimodal reasoning and action.
\newblock \emph{ArXiv}, abs/2303.11381.

\bibitem[{Yao et~al.(2023)Yao, Zhao, Yu, Du, Shafran, Narasimhan, and Cao}]{react}
Shunyu Yao, Jeffrey Zhao, Dian Yu, Nan Du, Izhak Shafran, Karthik~R Narasimhan, and Yuan Cao. 2023.
\newblock React: Synergizing reasoning and acting in language models.
\newblock In \emph{ICLR}.

\bibitem[{Yin et~al.(2023)Yin, Brahman, Ravichander, Chandu, Chang, Choi, and Lin}]{yin2023lumos}
Da~Yin, Faeze Brahman, Abhilasha Ravichander, Khyathi Chandu, Kai-Wei Chang, Yejin Choi, and Bill~Yuchen Lin. 2023.
\newblock Lumos: Learning agents with unified data, modular design, and open-source llms.
\newblock \emph{arXiv preprint arXiv:2311.05657}.

\bibitem[{Zhang et~al.(2023)Zhang, Hou, Xie, Sun, McAuley, Zhao, Lin, and Wen}]{zhang2023agentcf}
Junjie Zhang, Yupeng Hou, Ruobing Xie, Wenqi Sun, Julian McAuley, Wayne~Xin Zhao, Leyu Lin, and Ji-Rong Wen. 2023.
\newblock Agentcf: Collaborative learning with autonomous language agents for recommender systems.
\newblock \emph{arXiv preprint arXiv:2310.09233}.

\bibitem[{Zhuang et~al.(2023)Zhuang, Yu, Wang, Sun, and Zhang}]{zhuang2023toolqa}
Yuchen Zhuang, Yue Yu, Kuan Wang, Haotian Sun, and Chao Zhang. 2023.
\newblock Toolqa: A dataset for llm question answering with external tools.
\newblock \emph{arXiv preprint arXiv:2306.13304}.

\end{thebibliography}

\bigskip
\appendix
\section{Appendix}~\label{appendix}

\subsection{Details of Action Distillation}\label{sec:app:distillation}

Our specialized action distillation (\distill) trains three student models separately using the task-solving trajectory of a powerful model, \ie GPT-4 in our implementation.
These three student models are trained to conduct specific actions of the \first, \second, and \third, respectively.
Their initial parameters weights $\theta$ are  initialized from the same open-source model $\mathcal{M}_{\theta}$.
Since we use LoRa~\cite{hu2021lora} for parameter-efficient tuning, the optimization objective of our distillation is to search for the delta parameter $\Delta\theta$ to minimize the loss function.
Here, we introduce their detailed optimization objectives.

\paragraph{Notations.}
As mentioned in \S~\ref{sec:method}, we denote an input task as $x$, which is solved in a step-by-step manner while the task-solving context is denoted as $\mathcal{H}$.
In $i$th step, the context $\mathcal{H}_i$ contains historical planning $t_{<i}$ and execution results $r_{<i}$.
The planning $t$ specifies a tool to use in a current step which is selected from a candidate toolset $\mathcal{S}$.

\paragraph{Training of \first.}
Given a task $x$, we train the \first \ga to decompose $x$ into simpler sub-tasks and ground each sub-task into tool-use planning $t$ on the condition of the current context $H$ and revise incorrect planning following the feedback $f_{R\rightarrow G}$ of the \third \oa.
For each step $t_i$, we use the standard language modeling loss for optimization, which can be formulated as:
\begin{equation*}
    \begin{aligned}
       \mathcal{L}_G
         =& - \log P_{\theta+\Delta\theta_{G}} \left(t_i^{j} | x,\mathcal{H}_i, \mathcal{S}, \{t_i^{<j}, f_{R\rightarrow G}^{<j}\} \right)  \\
   \end{aligned}
    \label{eq:loss1}
\end{equation*}
Here, the $j$ indicate the $j$th interaction between the agent \ga and \oa.
The $\{t_i^{<j}, f_{R\rightarrow G}^{<j}\}$ indicates the planning-review alternated from agent \ga to \oa.
The LoRa parameter of agent \ga is denoted as $\Delta\theta_{G}$.

\paragraph{Training of \second.}
Similarly, in the $i$th step, we train the \second \ea to execute a tool following the planning $t_i$ by generating an executable program, and then calibrate incorrect code following the review of agent \oa.
Formally, the optimization objective can be formulated as:
\begin{equation*}
    \begin{aligned}
       \mathcal{L}_E
         = & - \log P_{\theta+\Delta\theta_{E}} \left(c_i^{j} | x, t, d, \{c_i^{<j}, f_{R\rightarrow E}^{<j}\} \right)  \\
   \end{aligned}
    \label{eq:loss2}
\end{equation*}
Here, $d$ indicates the tool documentation.
The LoRa parameter of agent \ea is denoted as $\Delta\theta_{E}$.

\paragraph{Training of \third.}
The \third agent is trained to provide reviews for agent \ea and \oa, calibrating their incorrect actions, \ie planning or execution.
Thus, the optimization objective can be formulated as:
\begin{equation*}
\begin{aligned}
       \mathcal{L}_R
         = & -  \sum_{j=1}^{\alpha} \log P_{\theta+\Delta\theta_{R}} \left(f_{R\rightarrow G}^{j} | x, S, t_i^{j-1}\right) -  \\
          &  \sum_{j=1}^{\beta} \log P_{\theta+\Delta\theta_{R}} \left(f_{R\rightarrow E}^{j} | x, d, c_i^{j-1},r_i^{j-1}
         \right)
   \end{aligned}
    \label{eq:loss3}
\end{equation*}
Here, the LoRa parameter of agent \oa is denoted as $\Delta\theta_{R}$.

\subsection{Heuristic Strategies for Data Selection}\label{sec:app:filter}

We employ the following heuristic methods to filter low-quality tasks in the original ToolBench:

\begin{itemize}
\item Each task in ToolBench is paired with a list of candidate tools. Generally, the more candidate tools there are, the more complex the task. Thus, we filter out tasks with fewer than 10 candidate tools to ensure the overall complexity of the sampled tasks.
\item To improve the quality of our training dataset, we remove tasks if their tools are not callable or deprecated.
\item We remove tasks if their tools lack the required documentation or if the documentation is less than 100 words in length.
\end{itemize}

\subsection{Datasets}\label{sec:app:dataset}

\paragraph{Experiment dataset}
 We conduct experiments on three commonly-used datasets with tool learning tasks, including: 
 \begin{itemize}
     \item RestBench~\citep{restgpt}: a high-quality human annotated dataset consisting of 54 tools about movie scenarios.
     \item   RestBench-Spotify~\citep{restgpt}: a dataset with 40 tools for music scenarios.
     \item  ToolBench~\citep{toolllm}: a dataset containing diverse real-world tools across various applications, which contains the simple tasks, \ie solving a task with one single tool, and complex tasks, \ie executing multiple tools in a logic order to solve a task.
 \end{itemize}
 
We conduct experiments on the full dataset of TMDB and Spotify.
Due to the intensive inference cost of LLMs-based agents, we randomly sample 117 cases as test sets from the complex tasks in Toolbench datasets to evaluate the performance of our cooperative agent framework in solving practical tasks.
We will release the sampled task for the transparency consideration.

\paragraph{Extend existing datasets.}

The original ToolBench benchmark only provides a step-by-step task-solving trajectory of GPT-3.5, which consists of both valid ground truth tools and irrelevant tools.
However, our evaluation involves computing the overlap between model-selected tools with ground truth tools.
Therefore, we repurpose the ToolBench to support our evaluation methods.
Specifically, for each task, we extract the tools in the original solution provided by ToolBench and only retain the relevant tools that are required for solving the task.
We invite three well-educated masters with relevant research backgrounds to implement this process. To guarantee annotation quality, we ask at least two annotators to annotate the same task repeatedly. If there is a discrepancy between the two annotators (i.e., two annotators give different answers), we ask a third annotator to recheck it. We hold regular meetings and pre-annotation tests to ensure that each expert undergoes detailed training to familiarize themselves with our annotation task. We will release these repurposed tasks to facilitate future research.

\subsection{Evaluation Metrics Details}\label{app:metrics}

\paragraph{Automatic evaluation.}
We mainly employ Success Rate and Correct Path Rate as two automatic evaluation metrics, following previous works ~\citep{gpt4tools,gao2023confucius}. The Success Rate (\textbf{Success\%}) computes the proportion of successful query completions. Specifically, when all the ground-truth tools are executed correctly, the Success Rate is set to 1; otherwise, it is set to 0. The Correct Path Rate (\textbf{Path\%}) computes the F1 score between the generated tool sequence and the ground-truth tool sequence.

\paragraph{Human evaluation}\label{sec:app:human}

We conduct a human evaluation on two metrics, including:
(1) Executability (\textbf{Exec}): whether the multiple tools are invoked in a correct logical order to complete the task;
and (2) Utility: whether the execution results of tools can be used to generate an answer.
We invite three well-educated volunteers to evaluate 30 cases randomly sampled from RestBench-TMDB and RestBench-Spotify datasets, respectively, with a three-scale rating.
Using a 3-point scale over a binary scale provides an option for the annotators to factor in their subjective interpretation of the extent of success or failure of a system’s response to satisfy a user’s request.
The instructions used in our human evaluation are summarized as follows.
\makebox[\linewidth]{\rule{\linewidth}{0.4pt}}
\textit{The evaluation guideline for our human evaluation.}
\begin{lstlisting}[basicstyle=\small\ttfamily, breaklines=true, breakindent=0em, commentstyle=\color{red!50!green!50!blue!50}, frame=shadowbox, rulesepcolor=\color{red!20!green!20!blue!20},numbers=none,literate={`}{\textasciigrave}1]

In this evaluation task, you are provided with some question-solution pairs. The question can be only solved by using real-world tools (or APIs). The solution is a sequential tool-use process, involving multi-step tool callings.

Your task is to rate the quality of the solution on a three scale based on the following two  metrics: 
1. Executability: Whether multiple tools are invoked in a correct logical order to complete the task.
2. Utility: Whether the model can observe the relevant values from lengthy execution results, incorporate them to predict the next action, and finally output a correct answer.

We also provide scoring criteria for your reference. Please adhere to our criteria since we will re-check the score you provide.
Now, read the following criteria and rate the provided question-solution pairs. Note that, you are encouraged to give us feedback and share any confusion you may have.

==Scoring Criteria==

1. For the Executability metric:
- Three points: Call all necessary tools correctly and solve the task. Allow for redundant tools or inference steps. 
- Two points: Not fully calling all necessary tools correctly, partially solving the task. 
- One point: Only some sub-steps are solved and the entire task is not completed. And there is a lot of redundancy or incorrect reasoning.

2. For the Utility metric:
- Three points: A majority of the execution results of the tools are correctly used to address the question (minor mistakes are allowed).
- Two points: Only part of the execution results of the tools are used. For example, in a question requiring finding an actor's highest-grossing film, the correct solution is to sequentially look at all the films the actor has appeared in, instead of just counting the top-k like top-5 or top-10.
- One point: Only a small part of the execution results of the tools are used, while other useful intermediates are ignored.

\end{lstlisting}

\begin{figure*}[t]
        \centering
	\includegraphics[width=1\linewidth]{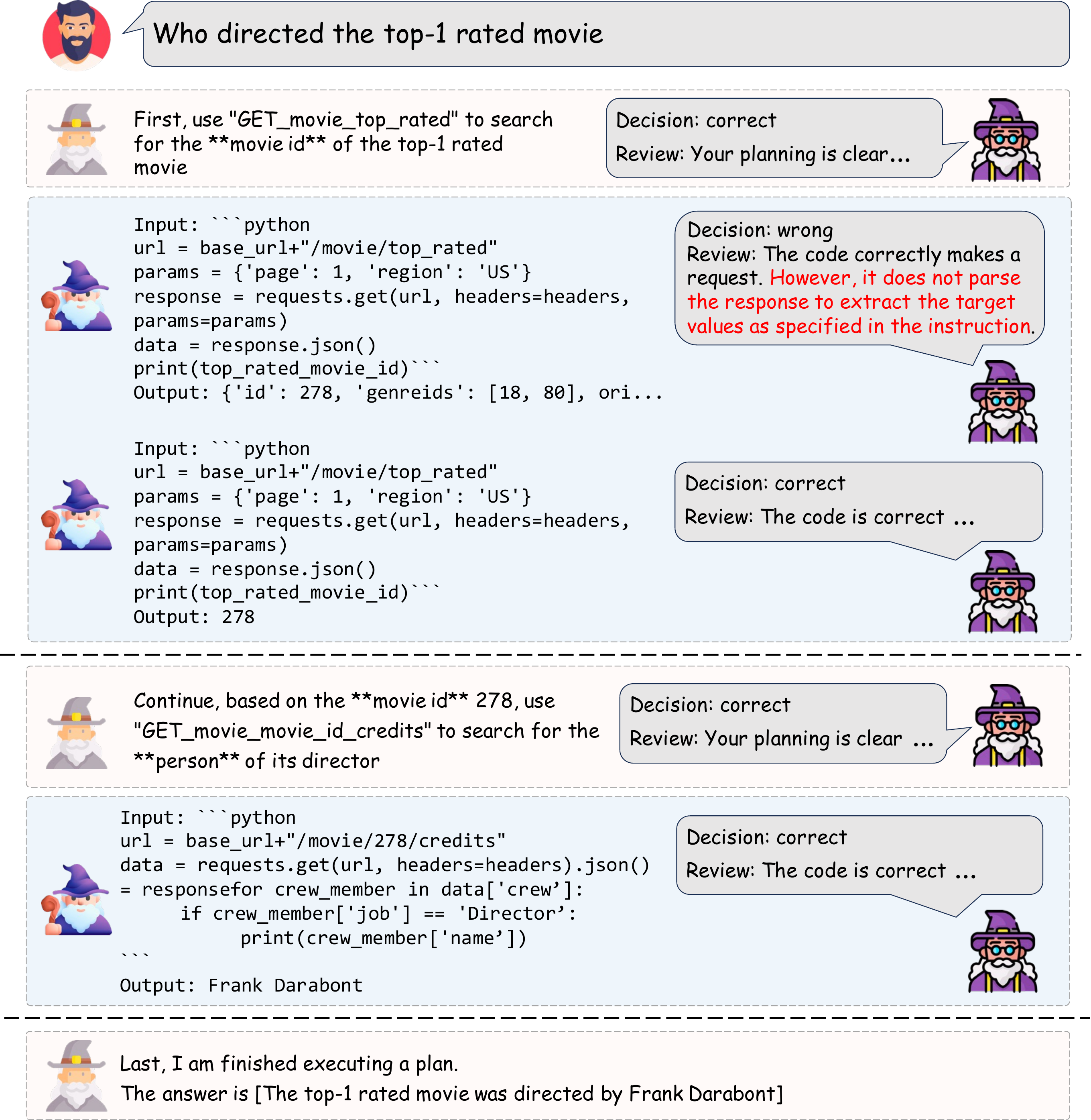}
        \caption{
An example to illustrate the proposed \textit{automatic agent communication} of our framework \ours.
For each turn, the communication starts with the planning-and-review between the \first and \third. 
Following the planning , the \second generates programs to execute tools and \daulg{calibrates the incorrect result with the review of \third}.
In this figure, we highlight the useful review of \third with 
\daulg{red}.
}
 \label{fig:case1}
\end{figure*}

\begin{figure*}[t]
        \centering
	\includegraphics[width=1\linewidth]{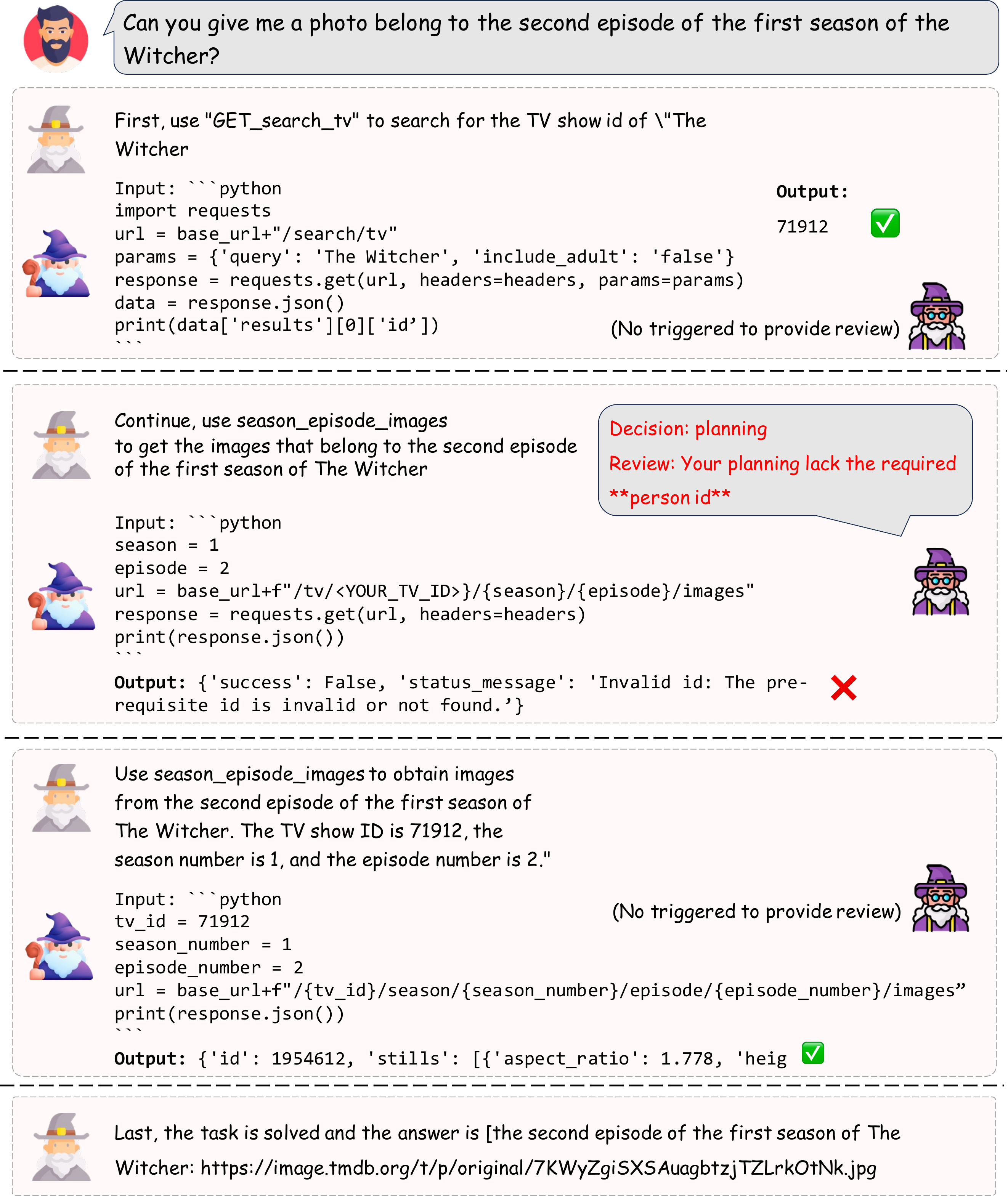}
        \caption{
An example to illustrate the proposed \textit{adaptive agent communication} in our framework \ours.
The agent flow mainly alternates from (1) generating tool-use planning by \first and (2) generating execution code by \second, in a step-by-step manner. The review agent is adaptively triggered to provide feedback \daulg{only when the generated code fails to execute correctly}.
In this figure, we highlight the review of \third with 
\daulg{red}.
}
 \label{fig:case2}
\end{figure*}

\subsection{Case Study}\label{sec:app:case}

We conduct several case studies and find that our method is effective at executing various tools and incorporating execution results to solve the input tasks.
Figure~\ref{fig:case1} presents a concrete example of the workflow of our proposed cooperative framework.




\paragraph{Case for our automatic agent communication.}

Figure~\ref{fig:case1} shows an example of our proposed \textit{automatic communication protocol}.
For each turn, the communication starts with the planning-and-review between the \first and \third. 
Following the planning , the \second generates programs to execute tools and calibrates the incorrect result with the review of \third.
For example, in the first turn, the agent \ga re-generate a planning following the review from agent \oa, and finally output a clear planning.
This example also illustrate the interaction between \first \ga and \third \oa, where the agent \ga calibrates its execution programs following the feedback of \oa,  and finally generate



\paragraph{Case for our adaptive agent communication}\label{sec:app:case-auto}
Figure~\ref{fig:case2} shows an example of our proposed \textit{adaptive communication protocol}. The agent flow mainly alternates between (1) generating tool-use planning by \first and (2) generating execution code by \second, in a step-by-step manner. The review agent is adaptively triggered to provide feedback only when the generated code fails to execute correctly.
For example, in the second turn, agent \ea initially generates a wrong program due to the lack of necessary arguments. Then, agent \oa reviews the current context, routes this error to agent \ga, and instructs \ga to supplement this argument, instead of directly shifting to the next state with an error response. This example intuitively illustrates the process of our adaptive interaction.




\end{document}